\documentclass{article}

\PassOptionsToPackage{numbers, compress}{natbib}

\usepackage[preprint]{neurips_2024}

\usepackage[utf8]{inputenc} 
\usepackage[T1]{fontenc}    
\usepackage{hyperref}       
\usepackage{url}            
\usepackage{booktabs}       
\usepackage{amsfonts}       
\usepackage{nicefrac}       
\usepackage{microtype}      
\usepackage{xcolor}         

\usepackage[pdftex]{graphicx}

\title{Quotient Network - A Network Similar to ResNet but Learning Quotients}

\author{%
	Peng Hui \\
	School of Agricultural Engineering\\
	Jiangsu University\\
	\texttt{2748477670@qq.com} \\
	\And
	Jiamuyang Zhao\\
	School of Agricultural Engineering\\
    Jiangsu University\\	
    \texttt{2222416076@stmail.ujs.edu.cn}
    \And
	Changxin Li \\
	School of Software Engineering \\
	Beihang University \\
	\texttt{793305696@qq.com}\\
	\And
	Qingzhen Zhu\\
	School of Agricultural Engineering \\
	Jiangsu University\\
	\texttt{qingzhen\_zhu@ujs.edu.cn} \\
}

\begin{document}
	
	\maketitle
	
	\begin{abstract}
		The emergence of ResNet provides a powerful tool for training extremely deep networks. The core idea behind it is to change the learning goals of the network. It no longer learns new features from scratch but learns the difference between the target and existing features. However, the difference between the two kinds of features does not have an independent and clear meaning, and the amount of learning is based on the absolute rather than the relative difference, which is sensitive to the size of existing features. We propose a new network that perfectly solves these two problems while still having the advantages of ResNet. Specifically, it chooses to learn the quotient of the target features with the existing features, so we call it the quotient network. In order to enable this network to learn successfully and achieve higher performance, we propose some design rules for this network so that it can be trained efficiently and achieve better performance than ResNet. Experiments on the CIFAR10, CIFAR100, and SVHN datasets prove that this network can stably achieve considerable improvements over ResNet by simply making tiny corresponding changes to the original ResNet network without adding new parameters.	
	\end{abstract}

	\section{Introduction}
	Convolutional neural networks have demonstrated strong performance in computer vision tasks\cite{krizhevsky2012imagenet,redmon2016you,  girshick2015fast, ronneberger2015u}. In their continuous performance improvements, depth is the key factor in success\cite{simonyan2014very, szegedy2015going}. In order to successfully train deeper networks, in addition to initialization regularization methods\cite{glorot2010understanding,  he2015delving, ioffe2015batch}, a landmark breakthrough that cannot be ignored is the ResNet method\cite{he2016deep}. It changes the learning target to the residual value through a shortcut so that the network does not need to learn new features from scratch but learns the difference between the new and old features, thereby reducing the learning difficulty. When the weight parameters are relatively small, it is easier to maintain the identity mapping so that a deep network is at least no worse than a shallow network.
	
	However, this also brings about two problems. The arithmetic difference between the new and old features is an absolute difference that does not fully utilize the size information of the old features. Obviously, for values 0.1 and 1, the effect of adding the same number 0.1 is quite different. Ignoring the size information of the old features will cause the learned intermediate values to be more sensitive to the size of the old features. The result is that the transformation is too strong for some old features and too weak for others. More importantly, CNN differs from RNN because its input and output feature types differ. Its different layers will learn low/medium/high-level features\cite{zeiler2014visualizing}. Therefore, the difference between the new and old features is not an intermediate feature with a very clear and independent meaning. That may make the functions to be learned by the network too complex and increase the learning difficulty. These two problems will eventually cause ResNet to be unable to utilize the performance of the deep network fully.
	
	For these reasons, we propose a more natural network to make learning easier. The network changes the learning goal to the quotient between the new and the old features, so the final feature is obtained by multiplying the old features by the quotient, so we call our network the quotient network. This network can perfectly solve these two problems of ResNet. It learns the relative difference (i.e., the quotient) between the new and old features, which better allows the network to take full advantage of the size information of the old features compared to ResNet, generating quantities insensitive to the size of the old features. That makes the quotient network more influential in transforming each old feature than ResNet. Looking back at the nature in which we live, we can see that the quotient of two different classes of features is more likely to be a third feature with an independent and clear meaning than the difference. For example, the value of mass divided by volume is more intuitively clear than the value of mass minus volume, the former is density; the value of force divided by mass is more meaningful than the value of force minus mass, the former is acceleration; the value of voltage divided by current is more straightforward than the value of voltage minus current, the former is resistance, and so on. In the quotient network, our learning goal is the quotient of different features, which tends to have a specific meaning. Introducing such prior knowledge can reduce the complexity of the function that is to be learned by the network.
		
	Moreover, our network also has ResNet's advantages. Instead of learning the target features from scratch, it learns them by building on top of the old features. By making the activation function of the last layer of the quotient module pass through the (0, 1) point, the quotient network can also maintain the identity mapping more easily. Therefore, we have reasons to believe that our network performance is better than ResNet.
		
	We propose some empirical guidelines for designing such networks, including finding better activation functions and placement of activation functions. Based on these criteria, we have obtained a network with powerful performance, which can stably obtain better performance than ResNet without changing the number of parameters of ResNet and only slightly increasing the amount of calculation. In experiments on CIFAR10\cite{krizhevsky2009learning}, CIFAR100\cite{krizhevsky2009learning}, and SVHN\cite{netzer2011reading} datasets, we only made corresponding slight modifications to ResNets with different numbers of layers and then stably achieved better performance than ResNets, demonstrating this network's ease of use and power. Furthermore, by visualizing the quotient feature maps, we justify our motivations. 
	
	\paragraph{Comparison with attention mechanisms}	
	In the design process of neural networks, multiplication has also been widely used in attention mechanisms. For example, SENet\cite{hu2018squeeze} and CBAM\cite{woo2018cbam} allow networks to focus on more valuable information by multiplying the weights of channels or spatial locations. In contrast, our method does not add weights to existing features but learns new and different features. Unlike other attentions, self-attention\cite{vaswani2017attention, dosovitskiy2020image} updates each feature by calculating correlations with other features. In contrast, our network has no Q, K, and V operations before generating features. In addition, when generating each new token, self-attention multiplies all old tokens with different weights and then adds them up. In contrast, our network will only perform one point-to-point multiplication operation on all old features. Let us look at self-attention from a perspective similar to that of the quotient network and ResNet, where the attentions of a particular token relative to other tokens are the weights of the neuron generating the new token.
	
	\section{Related work}
	
	\paragraph{Branches}	
	What is developing simultaneously with the increasing depth and width of the network is the concept of branches. Inception\cite{szegedy2015going, ioffe2015batch, szegedy2016rethinking, szegedy2017inception} uses branches to concatenate the features of filters of different sizes. Densenet\cite{huang2017densely} connects each layer to every other layer to enrich features. ResNext\cite{xie2017aggregated} calculates more channel information by grouping different channels without increasing the amount of calculations and parameters. In object detection or semantic/instance segmentation, branches are also used to enrich feature information. FPN\cite{lin2017feature} uses branches to fuse the lower position information with higher semantic information of the network. UNet\cite{ronneberger2015u} supplements the detailed information of the image through the horizontal branches of the U-shaped network. Branches are also used to reduce the amount of calculations and parameters. MobileNet\cite{howard2017mobilenets} lightens the network by group convolutions in which the number of channels equals the number of groups, and ShuffleNet\cite{zhang2018shufflenet} further groups 1x1 convolutions through shuffle. ResNet\cite{he2016deep} is different from the above. It uses branches to change the objective function of network learning. Its unique perspective has achieved great success, making residual learning a widely used operation today\cite{chollet2017xception, sandler2018mobilenetv2, vaswani2017attention}.
	
	\paragraph{Gates and attention mechanisms}
	
	Multiplication is widely used in gates and attention mechanisms. In RNN, gates solve the long-distance dependency problem by controlling the flow of information\cite{hochreiter1997long, cho2014learning}. The attention mechanisms allocate limited computing resources to more valuable feature areas by multiplying the weights. SENet\cite{hu2018squeeze} allocates attention to channels through squeeze and excitation operations. Based on this, there are improvements to the pooling operation used to extract features in the squeeze process\cite{gao2019global, qin2021fcanet} and improvements to the fully connected method of excitation\cite{wang2020eca}. CBAM\cite{woo2018cbam} uses average and maximum pooling to allocate attention to channels and spaces. Unlike CBAM, which does channel first and then spatial attention, BAM\cite{park2018bam} adopts a parallel method of channels and spaces. A breakthrough achievement in attention mechanisms is the proposal of the transformer\cite{vaswani2017attention}, which was initially used in NLP. ViT\cite{dosovitskiy2020image} splits the image into some patches for encoding and then introduces the transformer into the field of visual tasks. However, the amount of data required is enormous, so DeiT\cite{touvron2021training} uses a distillation token to reduce the need for massive data. There are also many improvements to architecture, the swin transformer\cite{liu2021swin} uses local windows and cross windows to transform, and the pyramid transformer\cite{wang2021pyramid} uses a shrinking pyramid to reduce the amount of calculation and produce high-resolution output, these networks can be used as the backbone of a variety of visual tasks. In object detection\cite{carion2020end, zhu2020deformable, li2022exploring} or semantic/instance segmentation\cite{zheng2021rethinking, chen2021transunet, xie2021segformer}, many transformer-based methods have achieved good performance.
	
	\section{Quotient network}
	
	\subsection{Reviewing residual learning}
	In order to solve the problem of the number of layers increasing but the training accuracy decreasing, ResNet changes the learning goal. Assume that the function a specific network block wants to learn is \(H(x)\). ResNet does not learn this goal directly from scratch but learns \(F(x) = H(x) - x\), so the network structure becomes \(H(x) = F(x) + x\). This operation reduces the network's learning difficulty. It helps to keep the information unchanged because compared to directly learning the identity mapping, this method can approximate the identity mapping as long as the parameters are small enough so that \(F(x)\) is close to 0. The ease of learning identity mapping ensures that the training loss of deep networks will not be greater than that of shallow networks.
	
	However, as discussed in the introduction, the arithmetic difference between different feature types is not an independent feature with clear meaning. Although nonlinear multi-layer networks can approximate complex functions, the increased complexity of the objective function caused by the difference without clear meaning will degrade network performance. Moreover, the arithmetic difference itself cannot make good use of the size information of the old features. The same increment will cause the smaller values to be over-updated while the larger values to be under-updated, which is detrimental to the learning of the network.
	
	\subsection{Quotient learning}
	\label{sec3_2}
\begin{figure}
		\centering
		\includegraphics[width=0.7\linewidth]{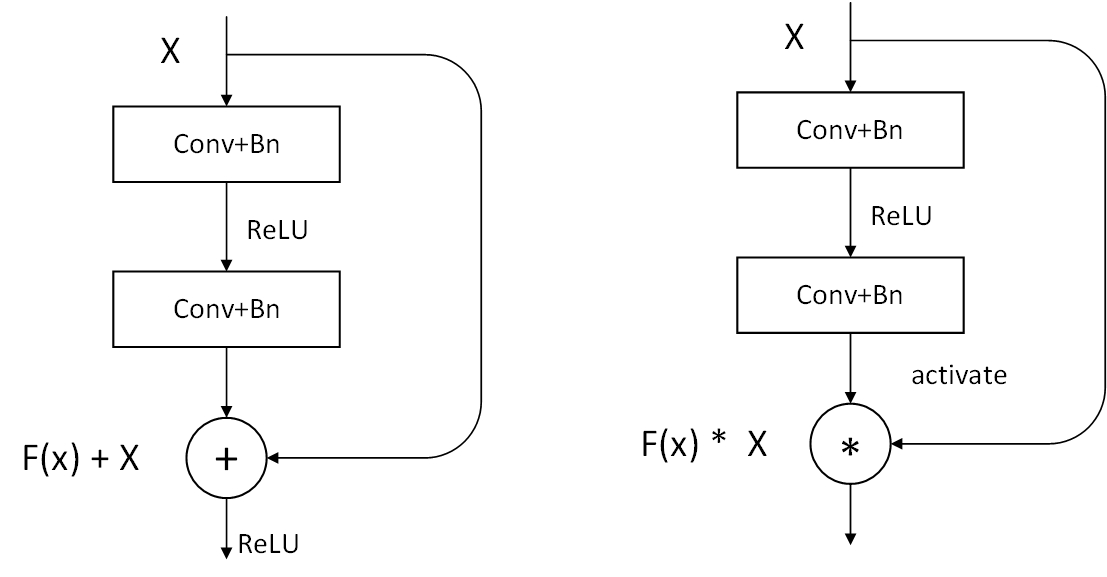}
		\caption{The residual module(left) and the quotient module(right)}
		\label{fig:fig1}
	\end{figure}

	We learn the quotient between two features to reduce the difficulty of network learning. As the introduction discusses, the transformation between two different features is often achieved by multiplying or dividing a third feature with an independent and clear meaning. As a result, the quotient we learn is more likely to be some meaningful feature, and this will reduce the complexity of the objective function. Moreover, through the multiplication operation, we can ensure that the same quotient value can enable old features of different sizes to be updated efficiently. Specifically, assuming that the function a particular network block wants to learn is \(H(x)\), we change its goal to \(F(x) = H(x) / x\), and the final network structure becomes \(H(x) = F(x) * x\). The comparison of ResNet with the quotient network is shown in Figure \ref{fig:fig1}. Unlike ResNet, our module is activated before the final multiplication.
	
	For the network to successfully learn the required quotient values, we need to design it specifically. Methods commonly used in convolutional networks may not be applicable to this network, such as the most commonly used activation function ReLU\cite{nair2010rectified}, which places half of the definition domain in the unsaturated zone and effectively solves the problem of vanishing gradients. However, if quotient values adopt ReLU, it may lead to exponential explosive growth of features, making the network unable to train. In Section \ref{sec3_3}, some empirical principles for selecting activation functions and model construction will be introduced.
	
	\subsection{The rules of designing}
	\label{sec3_3}
	After many attempts and failures, in which we constantly analyzed the reasons and made corrections, we finally propose the following empirical principles of quotient network design and explain their possible causes.
	
	\subsubsection{Choice of activation function for quotients}
	
	\textbf{The value range should not be too large or too small and should avoid negative numbers.} If it is too large, it will lead to the explosive growth of features due to continuous multiplication, and eventually, the network cannot be successfully trained, and the output will be white noise. If it is too small, the range of features that can be updated each time will be too small, affecting network performance. Moreover, the value range of the function should remain in the positive range because, for general activation functions(e.g., ReLU, Sigmoid), useful feature representations are often positive numbers. If multiplied by a negative number, this structure will be destroyed. If wanting to enlarge or reduce the features, multiplying by a positive number is more straightforward and intuitive.
	
	\textbf{The function should pass through the (0,1) point.} Like ResNet, we want to make it easier for the network to learn the identity mapping. When the weight parameters of the network are small, the weighted sum tends to be 0. At this time, ensuring that the value after activation is close to 1 helps to keep the previous features unchanged when multiplied by the previous features.
	
	\textbf{The function should be globally differentiable.} Because the value range of the activation function cannot be too large, that is, the value range is bounded, zero gradients cannot be used like ReLU in areas where the value range is close to the upper and lower bounds. Half of the domain in ReLU is in the unsaturated area. However, most domains of this activation function are the regions where the function values are close to the upper and lower bounds. If the gradient is completely zero, it will cause much performance loss.
	
	Formula \ref{eq1}, as the activation function, can perfectly meet the above requirements. It can ensure that the function is positive, passes through the (0,1) point, is globally differentiable, and can control the value range through \(\alpha\). In the appendix, we will compare the experimental results of different activation functions to illustrate the effectiveness of the design principles and the fact that the network using the activation function of Formula \ref{eq1} can indeed show good performance.
	
	\begin{equation}
		\label{eq1}
		activate(x) = sigmoid(x - ln(\alpha - 1)) * \alpha 
	\end{equation}
	
	\subsubsection{How to change the number of channels}
	In ResNet, when the number of channels increases, the number of channels of the old features does not match the number of channels of the learned residual features and cannot be added. The authors designed three methods to increase the number of channels of the original features. The first one is to add new channels directly to the old features, and the values in the newly added channels are all 0. The second one is to increase the number of channels of the old features by convolutional transformation when the number of channels increases. The third is to convolve the old features regardless of whether the number of channels is increased and then add the residuals to the old feature.
	
	In our network, because we use multiplication, if we multiply a channel with all zeros, the results of its subsequent operations will always be 0, and no useful information can be obtained. In the third, the amount of calculations is large. Therefore, like ResNet generally adopts the second method, we also adopt the second method of increasing channels.
	
	\subsubsection{Activation functions for other areas of the network}	
\begin{figure}
		\centering
		\includegraphics[width=0.3\linewidth]{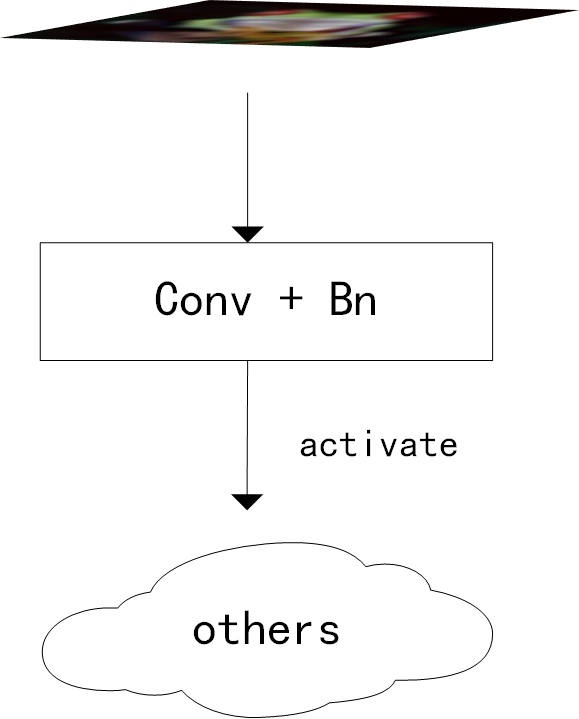}
		\caption{Convolution processing before stacking quotient modules}
		\label{fig:fig2}
	\end{figure}

\begin{figure}
		\centering
		\includegraphics[width=0.7\linewidth]{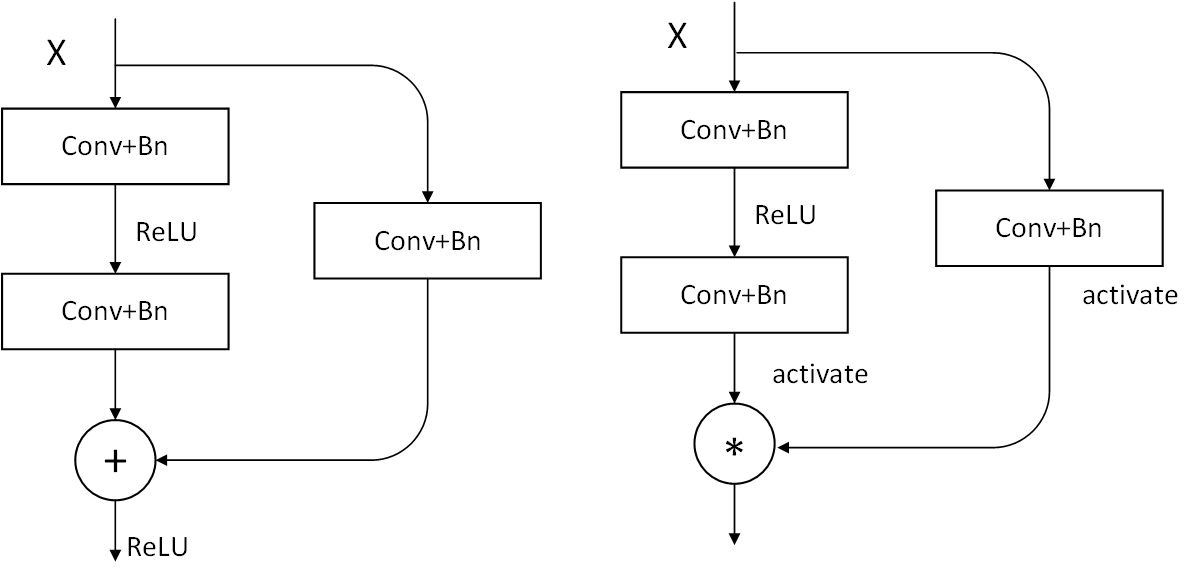}
		\caption{The residual module(left) and the quotient module(right) when changing the number of channels}
		\label{fig:fig3}
	\end{figure}

	In addition to the activation function of the last convolution before multiplication that needs to be specially designed, there are two other places worth noting. Before stacking quotient modules, the network needs to convolve the 3-channel RGB image to generate more channel features. Here, we find that it is better to use the same activation function as the last layer of the quotient module, as shown in Figure \ref{fig:fig2}. That will avoid excessively large output values when using ReLU and help keep the features consistent in size. For the same reason, when channels increase, this activation function should also be used after the old features are convolved to increase the number of channels, as shown in Figure \ref{fig:fig3}.
	
	\subsection{An example}
	\label{sec3_4}
	As an example, this section will present the simple residual networks used in the experiments on CIFAR10 and then give the corresponding quotient networks version. We use the network similar to the model the ResNet paper\cite{he2016deep} used in the CIFAR10 experiment as the basis. That is, the first layer of the network is a 3x3 convolution operation with stride one and output channels 16, and then three stages. Each stage consists of the same number of two layers' residual modules (both 3x3 convolutions) stacked. The first convolution of the first residual module of stage 2 and stage 3 adopts a 3x3 convolution with stride two and double channels. Therefore, the number of channels in the three stages is 16, 32, and 64, respectively, and the feature map size is 32, 16, and 8, respectively. Finally, after a global average pooling, the 10-way fully connected layer outputs prediction results. Depending on the number of modules stacked in each stage, networks of different depths can be formed, such as 20, 32, 44, 56, and 110 (the number of modules stacked in every stage is 3, 5, 7, 9, 18). Unlike the original paper shortcuts, which add new channels with all zeros when the number of channels increases, we use a 3x3 convolution with stride 2 to increase the number of channels.
	
	We modify the above network to obtain the corresponding quotient network. First, all residual modules in the three stages should be replaced, as shown in Figure \ref{fig:fig1}. Moreover, unlike ResNet, which all uses ReLU activation functions, we replace the activation functions in the first layer of the network and the convolutions that increase the number of channels in the shortcuts, as shown in Figure \ref{fig:fig2} and Figure \ref{fig:fig3}. 
	
	\subsection{The limitation and complex analysis}
	Frankly speaking, our network has increased the calculations by a certain amount. The increased amount of calculations comes from two aspects. The first is point-by-point multiplication. Since multiplication is implemented by many additions, compared with point-by-point addition, it will increase the calculations by a certain amount. The second is that our designed activation function will also increase the calculations' amount compared to the ReLU activation function. These will eventually cause our network to have a longer training and prediction time than ResNet with the same architecture. After our measurement, with the training mini-batch of 128 images, using the 56-layer model to train one CIFAR10 epoch, we need 22.603 s, ResNet needs 21.966 s, and to predict 128 pictures, we need 41.6 ms, ResNet needs 40.7 ms(all performed on a single RTX 4090 GPU).
	
	However, from the perspective of the parameters' amount, we do not add any parameters compared to ResNet. Moreover, as shown in Table \ref{table-6}, the accuracy of our 56-layer network is not only higher than the 56-layer ResNet but also higher than the 110-layer ResNet. After measuring the 110-layer ResNet, it takes 29.553s to train one epoch and 45.5 ms to predict 128 pictures, so the time consumption is much higher than that of our 56-layer network.
	
	\section{Experiments}
	We empirically demonstrate the effectiveness of the quotient networks on different datasets and compare the ResNet models to illustrate a steady and considerable improvement in our network performance compared to ResNets. Finally, we visualize the learned feature maps to justify the motivations of quotient networks.
	
	\subsection{Datasets}
	
	\paragraph{CIFAR10/CIFAR100}
	Both two datasets are composed of 32x32 RGB images, of which CIFAR10 has a total of 10 categories, and CIFAR100 has a total of 100 categories. The two datasets have the same number of training and test sets, where the training set consists of 50,000 images, and the test set consists of 10,000 images. We randomly select 5,000 images from the training set as the validation set, and the remaining 45,000 are used for training. Finally, we report the results of the test set. For data augmentation, we perform a random horizontal flip of the image, fill all sides with 4 pixels, and then randomly crop a 32x32 image. Finally, we normalize the data using channel means and standard deviations. For testing, we only evaluate the single view of the original 32x32 image  (including subtracted by the mean and divided by the std).
	
	\paragraph{SVHN}
	The street view house numbers (SVHN) dataset consists of 32x32 images, with 73,257 as the training data set and 26,032 as the test data set. We randomly select 6,000 images from the training set as the validation set, train on the remaining 67,257 images, and then test on the test set and report the results. In data preprocessing, we normalize the data using the channel means and standard deviations.

	\subsection{Training}
	We follow the training method in the ResNet paper \cite{he2016deep}. All the networks are trained with stochastic gradient descent (SGD) on all three datasets, with a momentum of 0.9, batch size of 128, and initial learning rate of 0.1. At the epoch 92 and 136,  the learning rate is divided by 10, and the final training epoch is 182.
	
	\subsection{Classification on CIFAR10}
	We compare our models with ResNets of different layers. Following our proposed network design rules, we choose Formula \ref{eq1} as the activation function. Moreover, this activation function is all used in the head and shortcuts(channels increasing). For the value of \(\alpha\), we experimentally found that the optimal value of \(\alpha\) is different for networks with different numbers of layers, and the optimal values of \(\alpha\) are 1.8, 1.7, and 1.5 for networks with 44, 56, and 110 layers, respectively, which are characterized by the fact that the larger the number of layers, the smaller the optimal value of \(\alpha\).
	
	 In order to make the experiments more credible, we conduct multiple experiments and report the statistical results. The experiment results are shown in Table \ref{table-1}. When comparing networks with the same number of layers, the accuracy of our network is always higher than that of the corresponding ResNet, and the difference is even larger when the number of layers increases. The accuracy of our 44-layer network is already close to that of the 56-layer ResNet, and when we increase the number of layers to 56, the accuracy of our network is even higher than that of the 110-layer ResNet. Thus, it can be proved that our network performs much better than ResNets.

	\begin{table}
		\caption{Comparison with ResNets of different layers on the CIFAR10 dataset. Results are expressed as "mean ± std".}
		\label{table-1}
		\centering	
		\resizebox{1.0\columnwidth}{!}{
			\begin{tabular}{lllll}
				\toprule
				& \multicolumn{2}{c}{quotient network} &
				\multicolumn{2}{c}{ResNet}   \\ 
				\cmidrule(r){2-3}     \cmidrule(r){4-5} 
				\#params             & network & accuracy(\%)                 & network   & accuracy(\%)       \\
				\midrule		
				0.68M & quotient network44  & 92.78±0.25  & Resnet44  & 92.61±0.33 \\
				0.87M & quotient network56 & 93.1±0.15           & Resnet56  & 92.84±0.18 \\
				1.75M & quotient network110 & 93.44±0.17           & Resnet110 & 93.02±0.33 \\
				\bottomrule		
		\end{tabular}}
	\end{table}

	\subsection{Classification on CIFAR100 and SVHN}
	In order to verify that our proposed network is suitable for a variety of datasets and not just showing higher accuracy on CIFAR10, we conduct experiments on both CIFAR100 and SVHN. Since the experiments aim to demonstrate that our network outperforms ResNet on multiple datasets rather than to improve the accuracy on these datasets, we do not conduct special designs. Specifically, on the SVHN dataset, we use the same network as on the CIFAR10 dataset. On the CIFAR100 dataset, we double the number of channels in the network so that the number of channels in the three stages is 32, 64, and 128, respectively, and replace the 10-way fully connected layer with a 100-way one. As a side note, the value of \(\alpha\) in the activation function is kept unchanged on both datasets, i.e., 1.8, 1.7, and 1.5 for the 44-, 56-, and 110-layer networks, respectively.
	
	As on CIFAR10, we conduct multiple experiments and report the statistical results. The results are shown in Table \ref{table-2} and Table \ref{table-3}. As can be seen, our networks stably outperform ResNets on both SVHN and CIFAR100 datasets. Although the quotient network and ResNet show some overfitting at the layer number of 110, our network still maintains a higher accuracy than ResNet. Moreover, our 44-layer network already outperforms the ResNets of all layer numbers. All these prove that our network has a general advantage over ResNet.

	\begin{table}
		\caption{Comparison with ResNets of different layers on the SVHN dataset. Results are expressed as "mean ± std".}
		\label{table-2}
		\centering	
		\resizebox{1.0\columnwidth}{!}{
			\begin{tabular}{lllll}
				\toprule
				& \multicolumn{2}{c}{quotient network} &
				\multicolumn{2}{c}{ResNet}   \\ 
				\cmidrule(r){2-3}     \cmidrule(r){4-5} 
				\#params & network & accuracy(\%)                 & network   & accuracy(\%)       \\
				\midrule
				0.68M & quotient network44 & 96.17±0.12         & Resnet44  & 95.98±0.04 \\
				0.87M & quotient network56 & \textbf{96.20±0.11}          & Resnet56  & 95.96±0.06 \\
				1.75M & quotient network110 & 96.12±0.05          & Resnet110 & \textbf{96.03±0.01} \\
				\bottomrule	
		\end{tabular}}
	\end{table}
	
	\begin{table}
		\caption{Comparison with ResNets of different layers on the CIFAR100 dataset. Results are expressed as "mean ± std".}
		\label{table-3}
		\centering	
		\resizebox{1.0\columnwidth}{!}{
			\begin{tabular}{lllll}
				\toprule
				& \multicolumn{2}{c}{quotient network} &
				\multicolumn{2}{c}{ResNet}   \\ 
				\cmidrule(r){2-3}     \cmidrule(r){4-5} 
				\#params & network & accuracy(\%)                 & network   & accuracy(\%)       \\
				\midrule
				2.72M & quotient network44 & 73.25±0.27           & Resnet44  & 72.66±1.24 \\
				3.50M & quotient network56  & \textbf{73.53±0.18}           & Resnet56  &
				\textbf{73.07±0.24} \\
				6.99M & quotient network110 & 73.00±0.55           & Resnet110 & 72.34±0.95 \\
				\bottomrule	
		\end{tabular}}
	\end{table}

	\subsection{Visualization}
	To verify the motivations proposed in the introduction, we visualize the intermediate feature (quotient for quotient network and residual for ResNet) maps calculated in the first three stacked modules of the 110-layer networks trained on CIFAR10. The feature maps when the input image is a frog are shown in Figure \ref{fig:fig4}, and the feature maps when the input images are other categories are shown in the Appendix. As can be seen from the figure, the quotient feature maps are clearer than the residual feature maps, and it is easier to see the complete structure of a frog from the quotient feature. This phenomenon is in line with our conjecture. The quotient of new and old features is more likely to be an independent and meaningful feature that can reflect a particular aspect of the characteristics of the frog. Therefore, it generates more clearly identifiable feature maps. On the contrary, ResNet learns the arithmetic difference of different types of features and lacks independent attribute meaning, so its feature maps are blurrier and more difficult to recognize. Moreover, our network learns relative difference (i.e., quotient), which is not sensitive to the size of old features, and it can exert a stable and effective influence on feature values of different sizes. That, in turn, makes our feature map information richer compared to ResNet's feature map information. It can even be seen that the feature maps of some  ResNet channels are approximately pure colors.
	
	\begin{figure}
		\centering
		\includegraphics[width=1\linewidth]{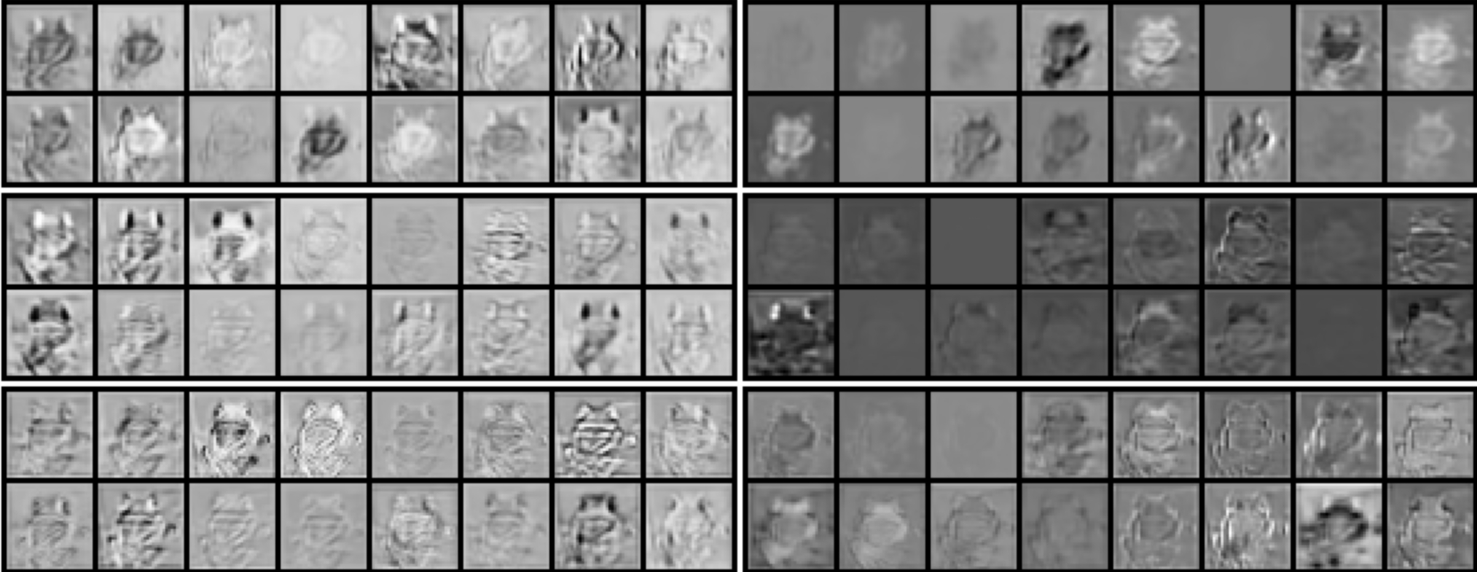}
		\caption{The middle feature maps when the input image is a frog. The left is for the quotient network, and the right is for ResNet. From top to bottom, it is for the first, second, and third stacked modules in order.}
		\label{fig:fig4}
	\end{figure}
	
	\section{Conclusion}
	We proposed a new network architecture called the quotient network. This kind of network changes the learning objective of a network block into the quotient of the target feature and the current feature. We presented several design guidelines for designing such a network and demonstrated the powerful performance of this network, which consistently outperforms the completely corresponding ResNet. Moreover, the design of this kind of network is straightforward and can be obtained by directly modifying ResNets.
	
	Due to time and hardware constraints, we did not use large-scale datasets such as ImageNet to learn and did not perform tasks such as object detection based on models pre-trained on ImageNet. Moreover, much can be studied in this network in the future. Since ResNet was proposed, many models have used residual learning (including transformers). Applying quotient learning to these models may also bring good performance or lead to some interesting problems. 
	
	{
		\small
		\bibliographystyle{IEEEtran}
		\bibliography{references}
	}
	
	\appendix
	\section*{\centering Appendix / supplemental material}
	In the Appendix, we first provide validation experiments for the quotient network design rules, as shown in Section \ref{appendix1}. We then offer intermediate feature visualizations when the input images are a bird, a cat, and a dog, as shown in Section \ref{appendix2}.
	
	\section{Validation experiments for design rules}
	\label{appendix1}
	We verify the design guidelines presented in Section \ref{sec3_3} of the main text by conducting experiments on the CIFAR10 dataset. Specifically, we compare the accuracy of using activation functions with different value ranges, whether the activation functions have negative values, whether the activation functions pass through the (0,1) point, whether the activation functions are globally differentiable, and whether the activation functions are used at the beginning of the network as well as in the shortcuts when increasing the number of channels.
	
	\paragraph{The value range of the activation function}
	We use 20-layer quotient networks as the basis for comparison. We use two classes of activation functions: the modified linear functions and the modified sigmoid functions. In order to keep the experimental results more comparable, we fix each activation function to pass through the (0,1) point, the convolutional layer at the beginning of the network, and the shortcuts when increasing the number of channels all use this activation function. The experimental results are shown in Table \ref{table-4}. As the table shows, for the 20-layer networks, the accuracy is the highest when the value range of the modified linear function is [0,4] or when the value range of the modified sigmoid function is (0,2). The accuracy will be reduced whether the value range is enlarged or reduced. Especially when the value range is infinite, training cannot be successful.

	\begin{table}[htbp]
		\caption{Using activation functions with different value ranges}
		\label{table-4}
		\centering
		\resizebox{1.0\columnwidth}{!}{
			\begin{tabular}{llllll}
				\toprule
				\multicolumn{3}{c}{mod linear} &
				\multicolumn{3}{c}{mod sigmoid}   \\ 
				\cmidrule(r){1-3}     \cmidrule(r){4-6} 
				activate function     & value range    & accuracy(\%) & activate function     & value range    & accuracy(\%) \\
				\midrule
				ReLU                  & \(\left [ 0, + \infty  \right )\)    & 10         &                     &              &            \\
				min(max(0, x+1), 8)   &  \(\left [ 0, 8 \right ]\)  & 88.32      & sigmoid(x - ln3) * 4 & (0,4)     & 91.15      \\
				min(max(0, x+1), 4.5) & \(\left [ 0, 4.5 \right ]\) & 90.75      & sigmoid(x - ln1.5) * 2.5 & (0,2.5)   & 91.57      \\
				min(max(0, x+1), 4)   & \(\left [ 0, 4 \right ]\)  & \textbf{91.01}      & sigmoid(x) * 2           & (0,2)     & \textbf{91.72}      \\
				min(max(0, x+1), 3.5) & \(\left [ 0, 3.5 \right ]\) & 90.98      & sigmoid(x - ln0.5) * 1.5 & (0,1.5)   & 91.44      \\
				min(max(0, x+1), 2)   & \(\left [ 0, 2 \right ]\)   & 90.71      &                          &           &     \\
				\bottomrule      
		\end{tabular}}
	\end{table}

    \paragraph{Whether to contain negative region}
    We continue to use 20-layer networks as the basis for comparison. For the modified linear functions, keep the value range size unchanged at 4; for the modified sigmoid, keep the value range size unchanged at 2. Keep each function passing through the (0,1) point. The convolution at the beginning of the network and the shortcuts when increasing the number of channels all use this activation function. We only change whether the value range includes the negative area and the size of the negative area, as shown in Table \ref{table-5}. Whether a modified linear function or a modified sigmoid, its accuracy will be reduced when its value range contains the negative area. It can be seen that the larger the area containing negative numbers, the greater the accuracy decrease.

	\begin{table}[htbp]
		\caption{Whether the activation functions have negative values}
		\label{table-5}
		\centering	
		\resizebox{1.0\columnwidth}{!}{
			\begin{tabular}{llllll}
				\toprule
				\multicolumn{3}{c}{mod linear} &
				\multicolumn{3}{c}{mod sigmoid}   \\ 
				\cmidrule(r){1-3}     \cmidrule(r){4-6} 
				activate function        & value range       & accuracy(\%)   & activate function          & value range   & accuracy(\%)   \\
				\midrule
				min(max(0, x+1), 4)      & \(\left [ 0, 4 \right ]\)       & \textbf{91.01} & sigmoid(x) * 2           & (0, 2)      & \textbf{91.72} \\
				min(max(-0.5, x+1), 3.5) & \(\left [ -0.5, 3.5 \right ]\) & 90.56          & sigmoid(x + ln3) * 2 - 0.5 & (-0.5, 1.5) & 91.21          \\
				min(max(-1, x+1), 3)     & \(\left [ -1, 3 \right ]\)    & 90.37          & sigmoid(x + ln9) * 2 - 0.8 & (-0.8, 1.2) & 91.06   \\
				\bottomrule         
		\end{tabular}}
	\end{table}

	\paragraph{Whether to pass the point (0, 1)}
	Like ResNet, whether it passes the (0,1) point is the key to whether the deep network can more easily maintain features. So, we use two different depths of 20 and 32 layers for comparison. Similarly, the modified linear function value range is kept as [0,4], the modified sigmoid function is kept as (0,2), and the convolution at the beginning of the network and the shortcuts when increasing the number of channels all use the designed activation function, only the function value at 0 is transformed. The comparison of 20-layer networks is shown in Table \ref{table-6}, and the comparison of 32-layer networks is shown in Table \ref{table-7}. When the independent variable is 0, whether the value is greater or less than 1, the accuracy will be reduced, and as the depth increases, the accuracy decrease will be more obvious. Here, we find that when the modified linear function is used in the 32-layer networks, regardless of whether it passes through (0,1), the accuracy will be reduced compared to the 20-layer networks. That may be because [0,4] is no longer suitable for 32-layer networks using the modified linear functions, but it is not important for our experimental purposes. 
	
	\begin{table}[htbp]
		\caption{Whether the activation functions pass through the (0,1) point (20-layer networks)}
		\label{table-6}
		\centering	
		\resizebox{1.0\columnwidth}{!}{
			\begin{tabular}{llllll}
				\toprule
				\multicolumn{3}{c}{mod linear} &
				\multicolumn{3}{c}{mod sigmoid}   \\ 
				\cmidrule(r){1-3}     \cmidrule(r){4-6} 
				activate function     & passing point & accuracy(\%)                 & activate function    & passing point & accuracy(\%) \\
				\midrule
				min(max(0, x+0.5), 4) & (0, 0.5)      & 90.07 & sigmoid(x – ln3) * 2 & (0, 0.5)      & 91.43        \\
				min(max(0, x+1), 4)   & (0, 1)        & \textbf{91.01} & sigmoid(x) * 2       & (0, 1)        & \textbf{91.72}        \\
				min(max(0, x+1.5), 4) & (0, 1.5)      & 90.58 & sigmoid(x + ln3) * 2 & (0, 1.5)      & 91.46 \\    
				\bottomrule       
		\end{tabular}}
	\end{table}
	
	\begin{table}[htbp]
		\caption{Whether the activation functions pass through the (0,1) point (32-layer networks)}
		\label{table-7}
		\centering	
		\resizebox{1.0\columnwidth}{!}{
			\begin{tabular}{llllll}
				\toprule
				\multicolumn{3}{c}{mod linear} &
				\multicolumn{3}{c}{mod sigmoid}   \\ 
				\cmidrule(r){1-3}     \cmidrule(r){4-6} 
				activate function     & Passing point & accuracy(\%)                 & activate function    & Passing point & accuracy(\%) \\
				\midrule		
				min(max(0, x+0.5), 4) & (0, 0.5)      & 82.84 & sigmoid(x – ln3) * 2 & (0, 0.5)      & 91.72        \\
				min(max(0, x+1), 4)   & (0, 1)        & \textbf{86.47} & sigmoid(x) * 2       & (0, 1)        & \textbf{92.51}        \\
				min(max(0, x+1.5), 4) & (0, 1.5)      & 85.84 & sigmoid(x + ln3) * 2 & (0, 1.5)      & 91.9   \\
				\bottomrule		     
		\end{tabular}}
	\end{table}
	
	\paragraph{Globally differentiable or not}
	With the above results, it is easy to realize that the accuracy of the modified linear function is always much lower than the modified sigmoid function. For the quotient network, the modified sigmoid function, which is a globally differentiable function, is more appropriate as the activation function.
	
	\paragraph{The head and shortcuts(channels increasing)}
	We finally compare the accuracy changes caused by whether the designed activation function is used in the first convolution and shortcuts when the number of channels increases. Here, 20-layer networks are used as the basis, and a modified linear function of [0,4] value range and a modified sigmoid function of (0,2) value range are taken, and both functions pass through the (0,1) point. Then, use the activation function for the head, use the activation function for both the head and shortcuts(channels increasing), and do not use it either for both places for comparison, as shown in Table \ref{table-8}. We can see that the accuracy is the worst when not using it in both places, and the accuracy is second when using it only in the head. The best is to use it in both places, and the accuracy is significantly improved.
	
	\begin{table}[htbp]
		\caption{Placing the designed activation function at different positions}
		\label{table-8}
		\centering
		\resizebox{1.0\columnwidth}{!}{
			\begin{tabular}{llllll}
				\toprule
				\multicolumn{3}{c}{mod linear} &
				\multicolumn{3}{c}{mod sigmoid}   \\ 
				\cmidrule(r){1-3}     \cmidrule(r){4-6} 
				activate function   & position         & accuracy(\%)                          & activate function & position         & accuracy(\%)   \\
				\midrule		
				min(max(0, x+1), 4) & null             & {\color[HTML]{000000} 89.15}          & sigmoid(x) * 2    & null             & 90.67          \\
				min(max(0, x+1), 4) & head             & {\color[HTML]{000000} 89.57}          & sigmoid(x) * 2    & head             & 90.93          \\
				min(max(0, x+1), 4) & head + shortcuts & {\color[HTML]{000000} \textbf{90.01}} & sigmoid(x) * 2    & head + shortcuts & \textbf{91.72} \\
				\bottomrule
		\end{tabular}}
	\end{table}
	
    \section{Supplementary visualization}  
    \label{appendix2}    
\begin{figure}[htbp]
    	\centering
    	\includegraphics[width=1\linewidth]{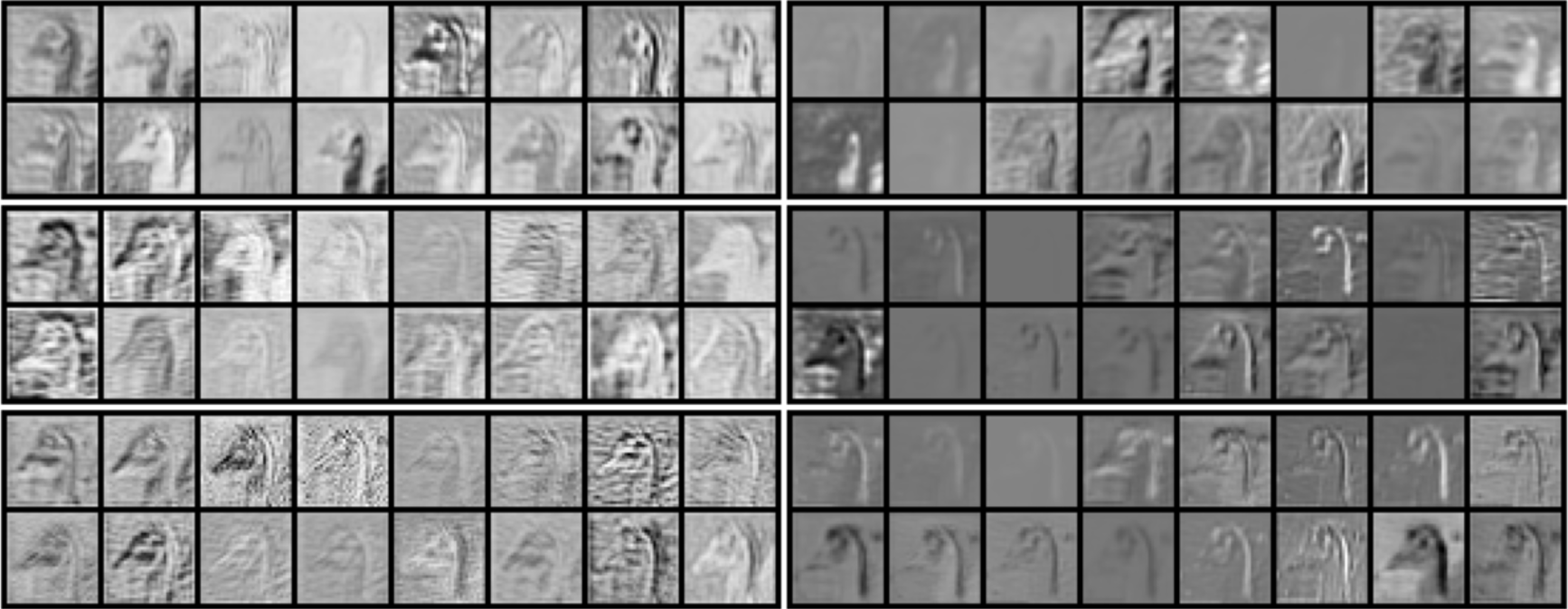}
    	\caption{The middle feature maps when the input image is a bird. The left is for the quotient network, and the right is for ResNet. From top to bottom, it is for the first, second, and third stacked modules in order.}
    	\label{fig:fig5}
    \end{figure}

\begin{figure}[htbp]
    	\centering
    	\includegraphics[width=1\linewidth]{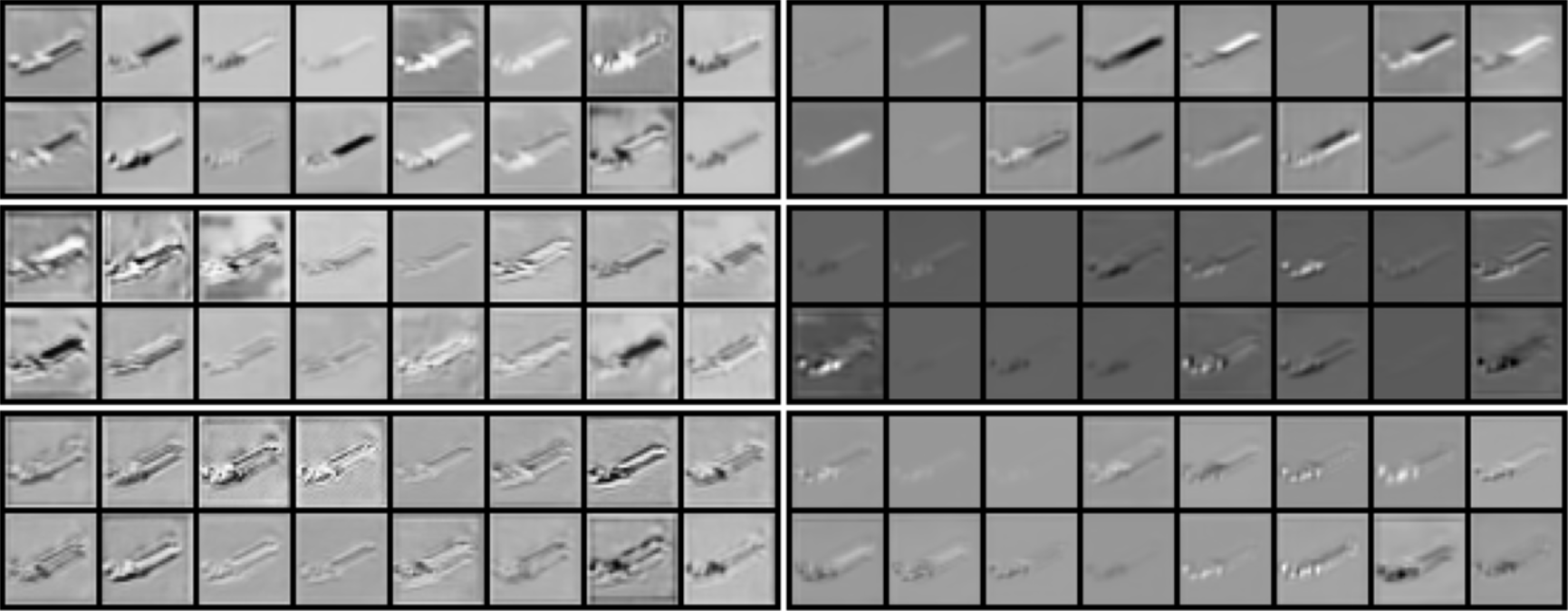}
    	\caption{The middle feature maps when the input image is a plane. The left is for the quotient network, and the right is for ResNet. From top to bottom, it is for the first, second, and third stacked modules in order.}
    	\label{fig:fig6}
    \end{figure}

\begin{figure}[htbp]
    	\centering
    	\includegraphics[width=1\linewidth]{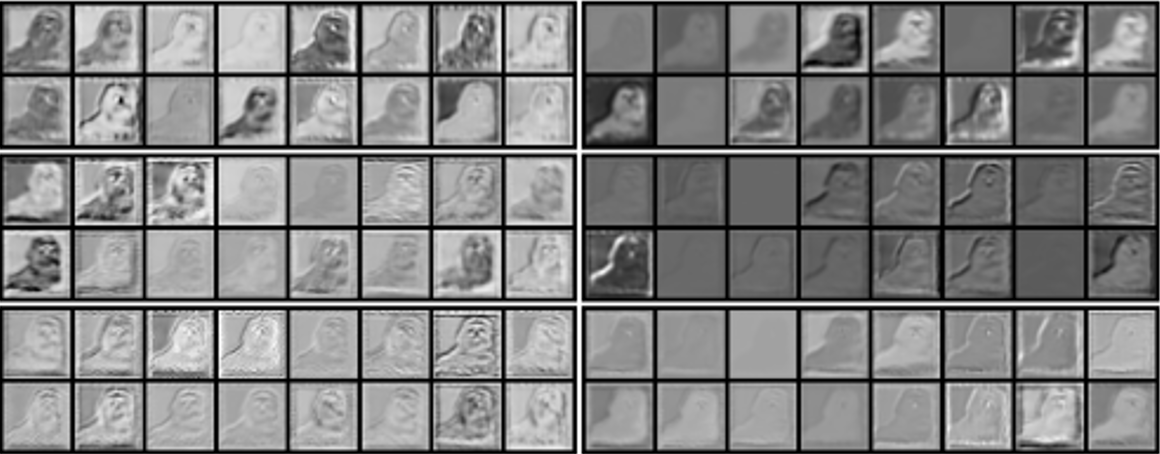}
    	\caption{The middle feature maps when the input image is a dog. The left is for the quotient network, and the right is for ResNet. From top to bottom, it is for the first, second, and third stacked modules in order.}
    	\label{fig:fig7}
    \end{figure}

\begin{figure}[htbp]
    	\centering
    	\includegraphics[width=1\linewidth]{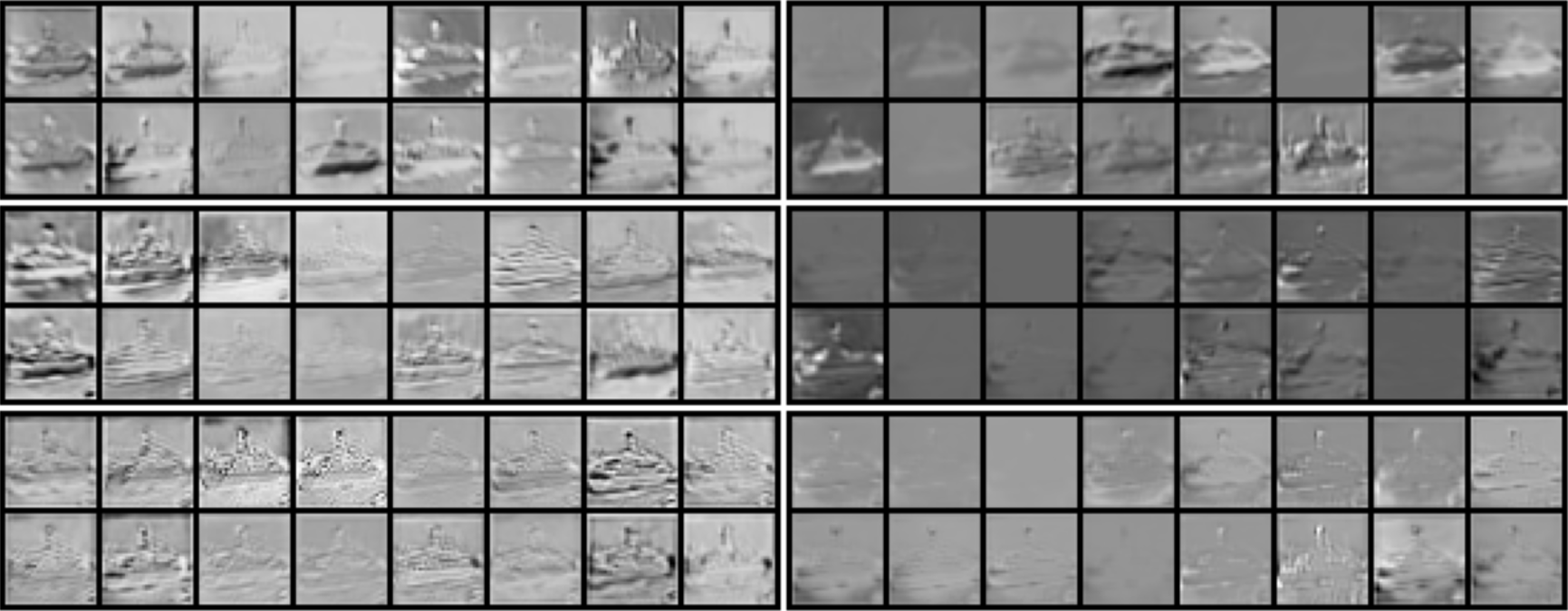}
    	\caption{The middle feature maps when the input image is a ship. The left is for the quotient network, and the right is for ResNet. From top to bottom, it is for the first, second, and third stacked modules in order.}
    	\label{fig:fig8}
    \end{figure}

\begin{figure}[htbp]
    	\centering
    	\includegraphics[width=1\linewidth]{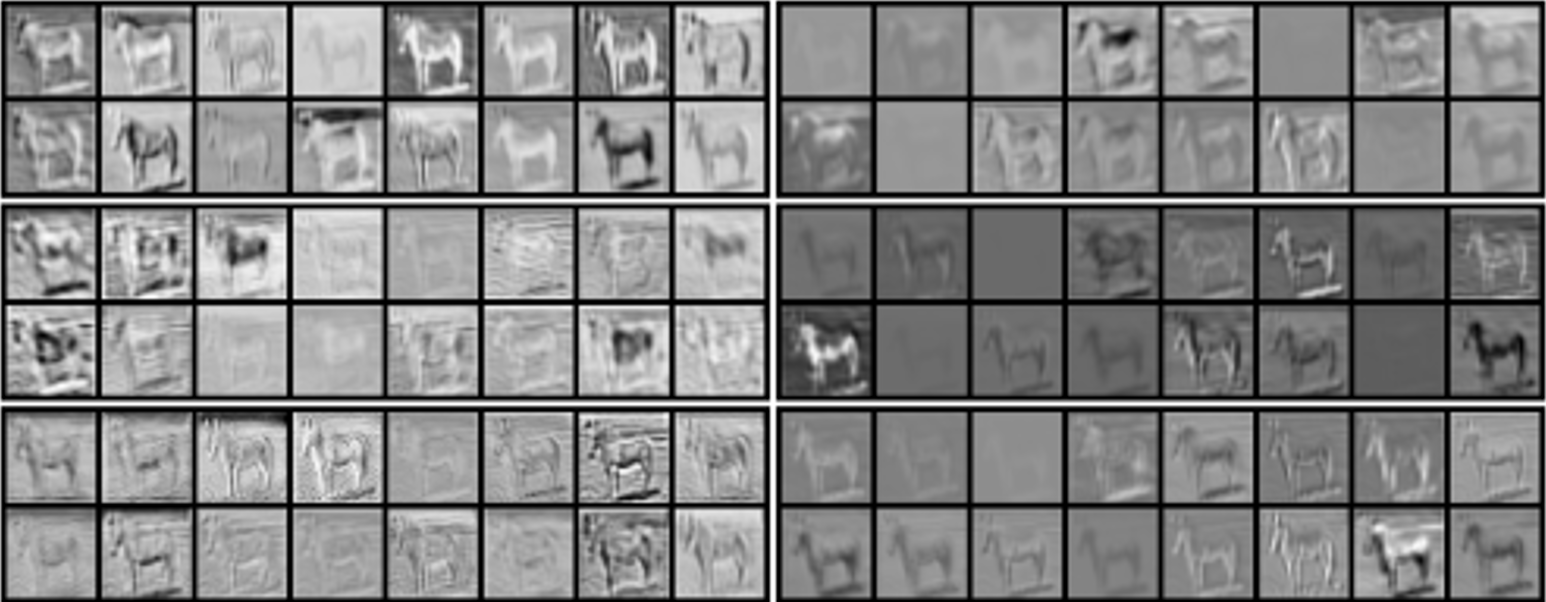}
    	\caption{The middle feature maps when the input image is a horse. The left is for the quotient network, and the right is for ResNet. From top to bottom, it is for the first, second, and third stacked modules in order.}
    	\label{fig:fig9}
    \end{figure}

    We visualize the first three intermediate feature (quotient for quotient network, residual for ResNet) maps when the inputs are pictures of other categories. Specifically, the bird in Figure \ref{fig:fig5}, the plane in Figure \ref{fig:fig6}, the dog in Figure \ref{fig:fig7}, the ship in Figure \ref{fig:fig8}, and the horse in Figure \ref{fig:fig9}. It can be seen that all of these pictures justify the motivations of the quotient network.
    
	\clearpage
	\section*{NeurIPS Paper Checklist}
	
	\begin{enumerate}
		
		\item {\bf Claims}
		\item[] Question: Do the main claims made in the abstract and introduction accurately reflect the paper's contributions and scope?
		\item[] Answer: \answerYes{} 
		\item[] Justification: In the abstract and introduction, we clearly present our motivation and the proposed network based on it, which is the contribution and scope of this paper. The specific network details and complete experimental validation will be presented later in the paper.
		\item[] Guidelines:
		\begin{itemize}
			\item The answer NA means that the abstract and introduction do not include the claims made in the paper.
			\item The abstract and/or introduction should clearly state the claims made, including the contributions made in the paper and important assumptions and limitations. A No or NA answer to this question will not be perceived well by the reviewers. 
			\item The claims made should match theoretical and experimental results, and reflect how much the results can be expected to generalize to other settings. 
			\item It is fine to include aspirational goals as motivation as long as it is clear that these goals are not attained by the paper. 
		\end{itemize}
		
		\item {\bf Limitations}
		\item[] Question: Does the paper discuss the limitations of the work performed by the authors?
		\item[] Answer: \answerYes{} 
		\item[] Justification: In Section 3.5, we analyzed our network's increased calculations compared to ResNet, which is the limitation.
		\item[] Guidelines:
		\begin{itemize}
			\item The answer NA means that the paper has no limitation while the answer No means that the paper has limitations, but those are not discussed in the paper. 
			\item The authors are encouraged to create a separate "Limitations" section in their paper.
			\item The paper should point out any strong assumptions and how robust the results are to violations of these assumptions (e.g., independence assumptions, noiseless settings, model well-specification, asymptotic approximations only holding locally). The authors should reflect on how these assumptions might be violated in practice and what the implications would be.
			\item The authors should reflect on the scope of the claims made, e.g., if the approach was only tested on a few datasets or with a few runs. In general, empirical results often depend on implicit assumptions, which should be articulated.
			\item The authors should reflect on the factors that influence the performance of the approach. For example, a facial recognition algorithm may perform poorly when image resolution is low or images are taken in low lighting. Or a speech-to-text system might not be used reliably to provide closed captions for online lectures because it fails to handle technical jargon.
			\item The authors should discuss the computational efficiency of the proposed algorithms and how they scale with dataset size.
			\item If applicable, the authors should discuss possible limitations of their approach to address problems of privacy and fairness.
			\item While the authors might fear that complete honesty about limitations might be used by reviewers as grounds for rejection, a worse outcome might be that reviewers discover limitations that aren't acknowledged in the paper. The authors should use their best judgment and recognize that individual actions in favor of transparency play an important role in developing norms that preserve the integrity of the community. Reviewers will be specifically instructed to not penalize honesty concerning limitations.
		\end{itemize}
		
		\item {\bf Theory Assumptions and Proofs}
		\item[] Question: For each theoretical result, does the paper provide the full set of assumptions and a complete (and correct) proof?
		\item[] Answer: \answerNA{} 
		\item[] Justification: Our network is proposed from an intuitive insight into residual learning, demonstrated through experimentation rather than theory.
		\item[] Guidelines:
		\begin{itemize}
			\item The answer NA means that the paper does not include theoretical results. 
			\item All the theorems, formulas, and proofs in the paper should be numbered and cross-referenced.
			\item All assumptions should be clearly stated or referenced in the statement of any theorems.
			\item The proofs can either appear in the main paper or the supplemental material, but if they appear in the supplemental material, the authors are encouraged to provide a short proof sketch to provide intuition. 
			\item Inversely, any informal proof provided in the core of the paper should be complemented by formal proofs provided in appendix or supplemental material.
			\item Theorems and Lemmas that the proof relies upon should be properly referenced. 
		\end{itemize}
		
		\item {\bf Experimental Result Reproducibility}
		\item[] Question: Does the paper fully disclose all the information needed to reproduce the main experimental results of the paper to the extent that it affects the main claims and/or conclusions of the paper (regardless of whether the code and data are provided or not)?
		\item[] Answer: \answerYes{} 
		\item[] Justification: In Section 3, we presented the complete details of the technique used in our proposed model. In Sections 3.4 and 4, we detailed the specific construction of the models used in the experiments, data processing, and the training and testing methodology.
		\item[] Guidelines: 
		\begin{itemize}
			\item The answer NA means that the paper does not include experiments.
			\item If the paper includes experiments, a No answer to this question will not be perceived well by the reviewers: Making the paper reproducible is important, regardless of whether the code and data are provided or not.
			\item If the contribution is a dataset and/or model, the authors should describe the steps taken to make their results reproducible or verifiable. 
			\item Depending on the contribution, reproducibility can be accomplished in various ways. For example, if the contribution is a novel architecture, describing the architecture fully might suffice, or if the contribution is a specific model and empirical evaluation, it may be necessary to either make it possible for others to replicate the model with the same dataset, or provide access to the model. In general. releasing code and data is often one good way to accomplish this, but reproducibility can also be provided via detailed instructions for how to replicate the results, access to a hosted model (e.g., in the case of a large language model), releasing of a model checkpoint, or other means that are appropriate to the research performed.
			\item While NeurIPS does not require releasing code, the conference does require all submissions to provide some reasonable avenue for reproducibility, which may depend on the nature of the contribution. For example
			\begin{enumerate}
				\item If the contribution is primarily a new algorithm, the paper should make it clear how to reproduce that algorithm.
				\item If the contribution is primarily a new model architecture, the paper should describe the architecture clearly and fully.
				\item If the contribution is a new model (e.g., a large language model), then there should either be a way to access this model for reproducing the results or a way to reproduce the model (e.g., with an open-source dataset or instructions for how to construct the dataset).
				\item We recognize that reproducibility may be tricky in some cases, in which case authors are welcome to describe the particular way they provide for reproducibility. In the case of closed-source models, it may be that access to the model is limited in some way (e.g., to registered users), but it should be possible for other researchers to have some path to reproducing or verifying the results.
			\end{enumerate}
		\end{itemize}

		\item {\bf Open access to data and code}
		\item[] Question: Does the paper provide open access to the data and code, with sufficient instructions to faithfully reproduce the main experimental results, as described in supplemental material?
		\item[] Answer: \answerYes{} 
		\item[] Justification: In the supplementary material, we provide the full experimental code and instructions.
		\item[] Guidelines:
		\begin{itemize}
			\item The answer NA means that paper does not include experiments requiring code.
			\item Please see the NeurIPS code and data submission guidelines (\url{https://nips.cc/public/guides/CodeSubmissionPolicy}) for more details.
			\item While we encourage the release of code and data, we understand that this might not be possible, so “No” is an acceptable answer. Papers cannot be rejected simply for not including code, unless this is central to the contribution (e.g., for a new open-source benchmark).
			\item The instructions should contain the exact command and environment needed to run to reproduce the results. See the NeurIPS code and data submission guidelines (\url{https://nips.cc/public/guides/CodeSubmissionPolicy}) for more details.
			\item The authors should provide instructions on data access and preparation, including how to access the raw data, preprocessed data, intermediate data, and generated data, etc.
			\item The authors should provide scripts to reproduce all experimental results for the new proposed method and baselines. If only a subset of experiments are reproducible, they should state which ones are omitted from the script and why.
			\item At submission time, to preserve anonymity, the authors should release anonymized versions (if applicable).
			\item Providing as much information as possible in supplemental material (appended to the paper) is recommended, but including URLs to data and code is permitted.
		\end{itemize}

		\item {\bf Experimental Setting/Details}
		\item[] Question: Does the paper specify all the training and test details (e.g., data splits, hyperparameters, how they were chosen, type of optimizer, etc.) necessary to understand the results?
		\item[] Answer: \answerYes{} 
		\item[] Justification: In Sections 3.4 and 4, we presented all the details of the models used in the experiments, of the methodology for data processing, training, and testing.
		\item[] Guidelines:
		\begin{itemize}
			\item The answer NA means that the paper does not include experiments.
			\item The experimental setting should be presented in the core of the paper to a level of detail that is necessary to appreciate the results and make sense of them.
			\item The full details can be provided either with the code, in appendix, or as supplemental material.
		\end{itemize}
		
		\item {\bf Experiment Statistical Significance}
		\item[] Question: Does the paper report error bars suitably and correctly defined or other appropriate information about the statistical significance of the experiments?
		\item[] Answer: \answerYes{} 
		\item[] Justification: In Section 4, we report the mean ± std for multiple replications of the experiment to make the experimental results more plausible.
		\item[] Guidelines:
		\begin{itemize}
			\item The answer NA means that the paper does not include experiments.
			\item The authors should answer "Yes" if the results are accompanied by error bars, confidence intervals, or statistical significance tests, at least for the experiments that support the main claims of the paper.
			\item The factors of variability that the error bars are capturing should be clearly stated (for example, train/test split, initialization, random drawing of some parameter, or overall run with given experimental conditions).
			\item The method for calculating the error bars should be explained (closed form formula, call to a library function, bootstrap, etc.)
			\item The assumptions made should be given (e.g., Normally distributed errors).
			\item It should be clear whether the error bar is the standard deviation or the standard error of the mean.
			\item It is OK to report 1-sigma error bars, but one should state it. The authors should preferably report a 2-sigma error bar than state that they have a 96\% CI, if the hypothesis of Normality of errors is not verified.
			\item For asymmetric distributions, the authors should be careful not to show in tables or figures symmetric error bars that would yield results that are out of range (e.g. negative error rates).
			\item If error bars are reported in tables or plots, The authors should explain in the text how they were calculated and reference the corresponding figures or tables in the text.
		\end{itemize}
		
		\item {\bf Experiments Compute Resources}
		\item[] Question: For each experiment, does the paper provide sufficient information on the computer resources (type of compute workers, memory, time of execution) needed to reproduce the experiments?
		\item[] Answer: \answerNo{} 
		\item[] Justification: Our experiments were conducted on multiple small datasets that did not require much computational resources, so we did not present computational resources in each experiment and only provided the computational resources for the experiments when analyzing computational complexity in Section 3.5.
		\item[] Guidelines:
		\begin{itemize}
			\item The answer NA means that the paper does not include experiments.
			\item The paper should indicate the type of compute workers CPU or GPU, internal cluster, or cloud provider, including relevant memory and storage.
			\item The paper should provide the amount of compute required for each of the individual experimental runs as well as estimate the total compute. 
			\item The paper should disclose whether the full research project required more compute than the experiments reported in the paper (e.g., preliminary or failed experiments that didn't make it into the paper). 
		\end{itemize}
		
		\item {\bf Code Of Ethics}
		\item[] Question: Does the research conducted in the paper conform, in every respect, with the NeurIPS Code of Ethics \url{https://neurips.cc/public/EthicsGuidelines}?
		\item[] Answer: \answerYes{} 
		\item[] Justification: Yes, we do.
		\item[] Guidelines:
		\begin{itemize}
			\item The answer NA means that the authors have not reviewed the NeurIPS Code of Ethics.
			\item If the authors answer No, they should explain the special circumstances that require a deviation from the Code of Ethics.
			\item The authors should make sure to preserve anonymity (e.g., if there is a special consideration due to laws or regulations in their jurisdiction).
		\end{itemize}

		\item {\bf Broader Impacts}
		\item[] Question: Does the paper discuss both potential positive societal impacts and negative societal impacts of the work performed?
		\item[] Answer: \answerNA{} 
		\item[] Justification: The foundational network model we proposed does not directly have social impacts.
		\item[] Guidelines:
		\begin{itemize}
			\item The answer NA means that there is no societal impact of the work performed.
			\item If the authors answer NA or No, they should explain why their work has no societal impact or why the paper does not address societal impact.
			\item Examples of negative societal impacts include potential malicious or unintended uses (e.g., disinformation, generating fake profiles, surveillance), fairness considerations (e.g., deployment of technologies that could make decisions that unfairly impact specific groups), privacy considerations, and security considerations.
			\item The conference expects that many papers will be foundational research and not tied to particular applications, let alone deployments. However, if there is a direct path to any negative applications, the authors should point it out. For example, it is legitimate to point out that an improvement in the quality of generative models could be used to generate deepfakes for disinformation. On the other hand, it is not needed to point out that a generic algorithm for optimizing neural networks could enable people to train models that generate Deepfakes faster.
			\item The authors should consider possible harms that could arise when the technology is being used as intended and functioning correctly, harms that could arise when the technology is being used as intended but gives incorrect results, and harms following from (intentional or unintentional) misuse of the technology.
			\item If there are negative societal impacts, the authors could also discuss possible mitigation strategies (e.g., gated release of models, providing defenses in addition to attacks, mechanisms for monitoring misuse, mechanisms to monitor how a system learns from feedback over time, improving the efficiency and accessibility of ML).
		\end{itemize}
		
		\item {\bf Safeguards}
		\item[] Question: Does the paper describe safeguards that have been put in place for responsible release of data or models that have a high risk for misuse (e.g., pretrained language models, image generators, or scraped datasets)?
		\item[] Answer: \answerNA{} 
		\item[] Justification: Our model poses no such risks.
		\item[] Guidelines:
		\begin{itemize}
			\item The answer NA means that the paper poses no such risks.
			\item Released models that have a high risk for misuse or dual-use should be released with necessary safeguards to allow for controlled use of the model, for example by requiring that users adhere to usage guidelines or restrictions to access the model or implementing safety filters. 
			\item Datasets that have been scraped from the Internet could pose safety risks. The authors should describe how they avoided releasing unsafe images.
			\item We recognize that providing effective safeguards is challenging, and many papers do not require this, but we encourage authors to take this into account and make a best faith effort.
		\end{itemize}
		
		\item {\bf Licenses for existing assets}
		\item[] Question: Are the creators or original owners of assets (e.g., code, data, models), used in the paper, properly credited and are the license and terms of use explicitly mentioned and properly respected?
		\item[] Answer: \answerYes{} 
		\item[] Justification: The paper references the data and models we used, and the license and terms of use are properly respected.
		\item[] Guidelines:
		\begin{itemize}
			\item The answer NA means that the paper does not use existing assets.
			\item The authors should cite the original paper that produced the code package or dataset.
			\item The authors should state which version of the asset is used and, if possible, include a URL.
			\item The name of the license (e.g., CC-BY 4.0) should be included for each asset.
			\item For scraped data from a particular source (e.g., website), the copyright and terms of service of that source should be provided.
			\item If assets are released, the license, copyright information, and terms of use in the package should be provided. For popular datasets, \url{paperswithcode.com/datasets} has curated licenses for some datasets. Their licensing guide can help determine the license of a dataset.
			\item For existing datasets that are re-packaged, both the original license and the license of the derived asset (if it has changed) should be provided.
			\item If this information is not available online, the authors are encouraged to reach out to the asset's creators.
		\end{itemize}
		
		\item {\bf New Assets}
		\item[] Question: Are new assets introduced in the paper well documented and is the documentation provided alongside the assets?
		\item[] Answer: \answerYes{} 
		\item[] Justification: New assets introduced in the paper are well documented, and the documentation is provided alongside the assets.
		\item[] Guidelines:
		\begin{itemize}
			\item The answer NA means that the paper does not release new assets.
			\item Researchers should communicate the details of the dataset/code/model as part of their submissions via structured templates. This includes details about training, license, limitations, etc. 
			\item The paper should discuss whether and how consent was obtained from people whose asset is used.
			\item At submission time, remember to anonymize your assets (if applicable). You can either create an anonymized URL or include an anonymized zip file.
		\end{itemize}
		
		\item {\bf Crowdsourcing and Research with Human Subjects}
		\item[] Question: For crowdsourcing experiments and research with human subjects, does the paper include the full text of instructions given to participants and screenshots, if applicable, as well as details about compensation (if any)? 
		\item[] Answer: \answerNA{} 
		\item[] Justification: The paper does not involve crowdsourcing and research with human subjects.
		\item[] Guidelines:
		\begin{itemize}
			\item The answer NA means that the paper does not involve crowdsourcing nor research with human subjects.
			\item Including this information in the supplemental material is fine, but if the main contribution of the paper involves human subjects, then as much detail as possible should be included in the main paper. 
			\item According to the NeurIPS Code of Ethics, workers involved in data collection, curation, or other labor should be paid at least the minimum wage in the country of the data collector. 
		\end{itemize}
		
		\item {\bf Institutional Review Board (IRB) Approvals or Equivalent for Research with Human Subjects}
		\item[] Question: Does the paper describe potential risks incurred by study participants, whether such risks were disclosed to the subjects, and whether Institutional Review Board (IRB) approvals (or an equivalent approval/review based on the requirements of your country or institution) were obtained?
		\item[] Answer: \answerNA{} 
		\item[] Justification: The paper does not involve crowdsourcing and research with human subjects.
		\item[] Guidelines:
		\begin{itemize}
			\item The answer NA means that the paper does not involve crowdsourcing nor research with human subjects.
			\item Depending on the country in which research is conducted, IRB approval (or equivalent) may be required for any human subjects research. If you obtained IRB approval, you should clearly state this in the paper. 
			\item We recognize that the procedures for this may vary significantly between institutions and locations, and we expect authors to adhere to the NeurIPS Code of Ethics and the guidelines for their institution. 
			\item For initial submissions, do not include any information that would break anonymity (if applicable), such as the institution conducting the review.
		\end{itemize}
		
	\end{enumerate}

\end{document}